\documentclass{article}
\usepackage{spconf,amsmath,epsfig,amssymb,multirow}
\usepackage{tabularx,ragged2e,booktabs,caption}

\begin{document}\sloppy

\def\x{{\mathbf x}}
\def\L{{\cal L}}

\title{CHEBYSHEV AND CONJUGATE GRADIENT FILTERS FOR GRAPH IMAGE DENOISING}
%
\name{Dong Tian, Hassan Mansour, Andrew Knyazev, Anthony Vetro}
\address{Mitsubishi Electric Research Labs (MERL)\\
        201 Broadway, 8th floor, Cambridge, MA 02139, USA\\
        \{tian; mansour; knyazev; avetro\}@merl.com}

\maketitle

\begin{abstract}
In 3D image/video acquisition, different views are often captured with varying noise levels across the views. In this paper, we propose a graph-based image enhancement technique that uses a higher quality view to enhance a degraded view. A depth map is utilized as auxiliary information to match the perspectives of the two views. Our method performs graph-based filtering of the noisy image by directly computing a projection of the image to be filtered onto a lower dimensional Krylov subspace of the graph Laplacian. We discuss two graph spectral denoising methods: first using Chebyshev polynomials, and second using iterations of the conjugate gradient algorithm. Our framework generalizes previously known polynomial graph filters, and we demonstrate through numerical simulations that our proposed technique produces subjectively cleaner images with about 1-3 dB improvement in PSNR over existing polynomial graph filters.
\end{abstract}
\begin{keywords}
3D images, image denoising, graph-based filtering, Krylov, Chebyshev, conjugate gradient
\end{keywords}
\section{Introduction}
\label{sec:intro}

In 3D video applications, multiple views are often presented at different quality levels due to various acquisition and compression conditions applied to the source signals. For example, changes of brightness or color may be produced by imaging sensors and circuitry of stereo cameras or even from shot noise. Traditional image enhancement techniques may be used to improve the quality of the degraded views. Moreover, depth map compression has become an integral part of emerging 3D video formats, e.g., 3D-AVC and 3D-HEVC. Hence, it is desirable to exploit depth information to enhance a low quality view.

Graph signal processing tools have recently been applied to classical image processing tasks \cite{shuman_signal_2013,Sunil_GlobalSip13,gadde2013bilateral,wang_icassp14,tian_icip2014}. For example, a typical interpolation problem was studied using spectral graph theory in \cite{Sunil_GlobalSip13}, where the upsampling problem was formulated as a regularized least squares problem. Later, this approach was extended to depth image upsampling in \cite{wang_icassp14}, and slight benefits of graph spectral domain processing were demonstrated. Recently, a similar graph based method to enhance noisy stereo images was investigated in \cite{tian_icip2014} utilizing the depth information to generate the guide image. However, these methods suffered from high complexity due to their requirement of computing a full eigenvalue decomposition on very high dimensional data.

An alternative approach that avoids full eigen-decompositions was adopted in \cite{gadde2013bilateral,GraphSP_HeatKernel:2008,WaveletsGraphs:2011} where the graph spectral filter was approximated by degree-$k$ polynomials. The filtering operation was then performed by applying the same polynomial as a function of the graph Laplacian matrix in the pixel domain. 

We follow the same line of thought and define a general framework for graph spectral filtering by directly computing the projection of the desired filtered signal onto an order-$(k+1)$ Krylov subspace for the graph Laplacian. We discuss how image denoising can be realized by low-pass filtering with a degree-$k$ Chebyshev polynomial with a predefined stop band. We then propose a parameter free filtering approach that utilizes $k$ iterations of the conjugate gradient (CG) algorithm to achieve the same denoising objective. We demonstrate through numerical experiments that our filters better suppress noise while maintaining image details compared to previous methods.

The remainder of the paper is organized as follows. Section 2 introduces the notation, gives a basic background on graph-based image processing, and reviews prior-art methods. Section 3 constructs a simple 4-connected graph using a guide image warped by 3D projection, and then proposes Chebyshev and CG filters to handle the image denoising problem. Section 4 describes our numerical experiments and analyses the results, demonstrating the improved performance of our filters. Section~5 concludes the work.

\section{Graph Based Image Processing}
\label{sec:bgbip}

\subsection{Basics of images on graphs}
\label{ssec:big}

In graph signal processing \cite{shuman_signal_2013}, an undirected graph $G=(V,E)$ consists of a collection of vertices, also called nodes, $V=\{1,2,\ldots,N\}$ connected by a set of edges $E=\{(i,j,w_{ij})\}$,$i,j\in V,$ where $(i,j,w_{ij})$ denotes an edge between nodes $i$ and $j$ associated with a weight $w_{ij}\geq 0$. A degree $d_{i}$ of a node $i$ is the sum of edge weights connected to the node $i$. For image processing applications, a pixel may be treated as a node in a graph, while the edge weight may be viewed as a measure of similarity of the pixels. 

An adjacency matrix $\mathbf{W}$ of the graph is a symmetric $N\times N$ matrix having entries $w_{ij}\geq 0$, and a diagonal degree matrix is $\mathbf{D}:=\textrm{diag}\{d_{1},d_{2},\ldots,d_{N}\}$. A graph Laplacian matrix $\mathbf{\mathbf{L}}:=\mathbf{D}-\mathbf{W}$ is a positive semi-definite matrix, 
thus admitting an eigendecomposition $\mathbf{L}=\mathbf{U}\Lambda\mathbf{U}^{T}$, where $\mathbf{U}$ is an orthogonal matrix with columns forming an orthonormal set of eigenvectors, and $\mathbf{\mathbf{\Lambda}}=\textrm{diag}\{\lambda_{1},\ldots,\lambda_{N}\}$ is a  matrix made of corresponding eigenvalues, all real. The smallest eigenvalue of the matrix $\mathbf{L}$ is always zero. 

Alternatively, a normalized symmetric Laplacian matrix is defined as $\mathbf{L_S}:=\mathbf{D}^{-1/2}\mathbf{L}\mathbf{D}^{-1/2}$, and is a positive semi-definite matrix. Hence, it admits an eigendecomposition $\mathbf{L_S}=\mathbf{U_S}\Lambda_S\mathbf{U_S}^{T}$, where $\mathbf{U_S}$ is an orthogonal matrix constructed from an orthonormal set of eigenvectors and $\mathbf{\mathbf{\Lambda_S}}$ is its corresponding diagonal eigenvalue matrix. The smallest eigenvalue of the matrix $\mathbf{L_S}$ is always zero, while the largest eigenvalue is conveniently bounded by two. 

The eigenvalues and eigenvectors of the Laplacian matrixes provide a spectral interpretation of graph signals, where the eigenvalues can be treated as graph Fourier frequencies, and the eigenvectors as generalized Fourier modes. 

A graph design can be associated to an underlying conventional image filter, traditionally used as one of the filter testing benchmarks. Specifically, for an input image $\mathbf{\hat{x}}_{in}$ the conventionally filtered output image $\mathbf{\hat{x}}_{out}$ can be written as,

%

\begin{equation}\label{eq:image_filter_matrix}
  \hat{\mathbf{x}}_{out}=\mathbf{D}^{-1}\mathbf{W}\hat{\mathbf{x}}_{in}=
\hat{\mathbf{x}}_{in}-\mathbf{D}^{-1}\mathbf{L}\hat{\mathbf{x}}_{in}.
\end{equation}

The original $\mathbf{L}$ or the symmetric normalized Laplacian $\mathbf{L_S}$ can serve as a foundation for image processing in the graph spectral domain. The peculiarity of the symmetric normalized Laplacian $\mathbf{L_S}$ is that it has all its eigenvalues located on the interval $[0,2]$ and that it requires pre- and post-processing of the images, i.e.,\ the signal in vertex domain needs to be normalized $\mathbf{x}_{in}=\mathbf{D}^{\frac{1}{2}}\hat{\mathbf{x}}_{in}$ before applying the filter, and denormalized back $\hat{\mathbf{x}}_{out}=\mathbf{D}^{-\frac{1}{2}} \mathbf{x}_{out}$ after being filtered. \vspace{0.1in}

\noindent \textbf{Remark:} In the rest of the paper, we assume, for certainty, that the symmetric normalized Laplacian $\mathbf{L_S}$ is always used, and we drop the subscript ``S'', denoting the symmetric normalized Laplacian as $\mathbf{L}=\mathbf{U}\Lambda\mathbf{U}^{T}$ and calling it simply the ``Laplacian'' hereafter. For example, conventional filter \eqref{eq:image_filter_matrix} in the new notation turns into  ${\mathbf{x}}_{out}={\mathbf{x}}_{in}-\mathbf{L}{\mathbf{x}}_{in}.$\vspace{0.1in}

\textit{Graph Spectral Filtering} (GSF) $\mathcal{H}$ can be designed for image processing purposes in the graph spectral domain, where $\mathcal{H}$ is a diagonal matrix, typically given as  $\mathcal{H}=h(\Lambda)$, where $h(\lambda)$ is a real valued function of a real variable  $\lambda$, determining the filter. 
The corresponding graph filter $\mathbf{H}$ in the vertex domain can be expressed as,
\begin{equation}
  \mathbf{H}=h(\mathbf{L})=\mathbf{U}\mathcal{H}\mathbf{U}^{T}.
\label{eq:graph_filter}
\end{equation}
For example, taking $h(\lambda)=\lambda$, we obtain the graph Laplacian $h(\mathbf{L})=\mathbf{L}$ itself that constitutes a high pass filtering operator attenuating low frequency coefficients and boosting high frequency coefficients of a signal.

In \cite{gadde2013bilateral} and \cite{tian_icip2014}, the filtered signal is computed by solving the following regularized least squares problem,
\begin{equation}
  \mathbf{x}^* = \arg\min\limits_{\mathbf{x} \in \mathbb{R}^N} \frac{1}{2}\|\mathbf{x} - \mathbf{b}\|^2  + \frac{\rho}{2}\|h_\text{R} \mathbf{x}\|^2,
\label{eq:classicGraphDenoising}
\end{equation}
where $\mathbf{b}:={\mathbf{x}}_{in}$ is the noisy signal, $h_\text{R}$ is a penalty  high pass graph filter, and $\rho$ is a regularization parameter. We refer to this formulation as graph based joint bilateral filter (GBJBF). The above problem admits an analytic solution of the form,
$\mathbf{x}^*  =  \mathbf{U} (\mathbf{I} + \rho \mathcal{H_R}^2 )^{-1} \mathbf{U}^t \mathbf{b} = (\mathbf{I} + \rho \mathbf{H}_\text{R}^2) ^{-1} \mathbf{b}$.
 
Choosing $\mathcal{H_R}=\Lambda$ results in the penalty filter $\mathbf{H}_\text{R} = \mathbf{L}$. Consequently, the GBJBF filter \eqref{eq:classicGraphDenoising} is given by,
\begin{equation}
  h_{\text{GBJBF}}(\lambda)=(1+\rho \lambda^2)^{-1}.
\label{eq:gbjbf_vertex}
\end{equation}
In \cite{gadde2013bilateral}, the filter function $h_{\text{GBJBF}}(\cdot)$ \eqref{eq:gbjbf_vertex} is approximated by a truncated degree-$k$ Chebyshev polynomial expansion, which we denote by $h_{\text{k-POLY}}(\cdot)$. 

We call a filter ``polynomial,'' if $h(\lambda)$ is a polynomial function in $\lambda$. For example, GBJBF filter $h_{\text{GBJBF}}(\cdot)$  \eqref{eq:gbjbf_vertex} is not polynomial, while its approximation $h_{\text{k-POLY}}(\cdot)$ is. The conventional filter \eqref{eq:image_filter_matrix} is given by $h(\lambda)=1-\lambda$, which is a first degree polynomial of $\lambda$, i.e.,\ is also polynomial. We note that application of the conventional filter via \eqref{eq:image_filter_matrix} does not require knowledge of the eigendecomposition of the Laplacian $\mathbf{L}$, involving instead only a calculation of the product of the matrix $\mathbf{L}$ and a given vector. 

This is a common feature of polynomial filters, including the filter given by $h_{\text{k-POLY}}(\cdot)$ \cite{gadde2013bilateral} , making polynomial filtering computationally attractive.  If the underlying polynomial is given via its roots or by a recursive formula, its action on an image can be implemented, using only the calculation of the product of the matrix $\mathbf{L}$ and a vector. 

\begin{figure}[t]
\begin{minipage}[b]{0.48\linewidth} 
  \centering
  \centerline{\includegraphics[width=2cm]{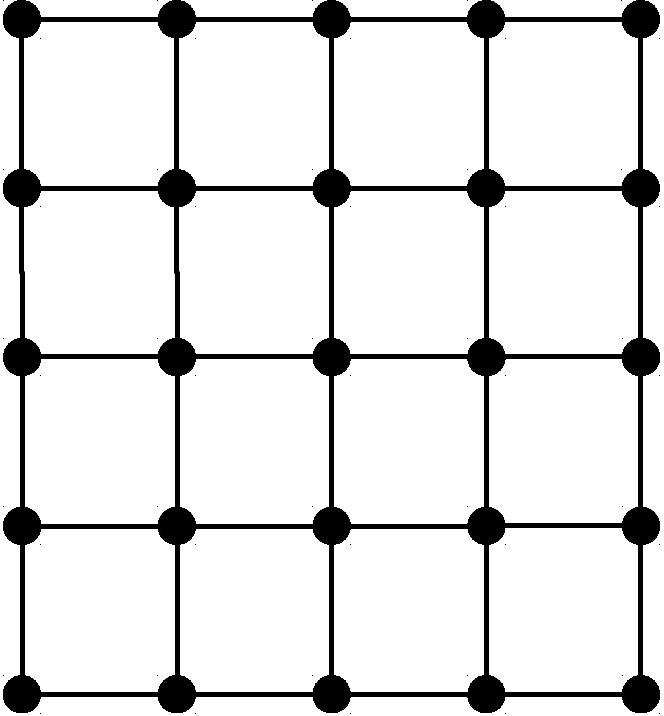}}
  \centerline{(a)}
\end{minipage}
\hfill
\begin{minipage}[b]{0.48\linewidth}
  \centering
  \centerline{\includegraphics[width=2cm]{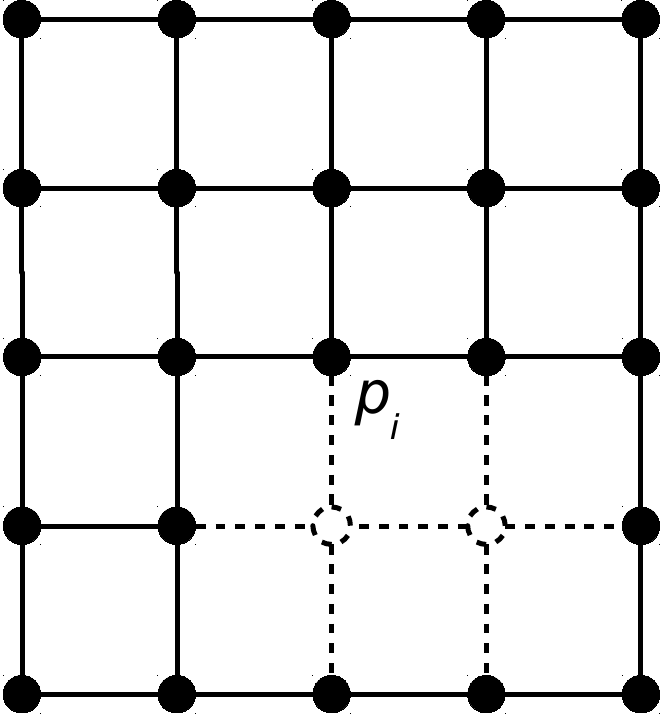}}
  \centerline{(b)}
\end{minipage}

\caption{Graph structure (a) used in \cite{leo_book}; (b) proposed in Section \ref{ssec:pgs} of this paper.}
\label{fig:graph}
\end{figure}

%
%

\subsection{Graph-based image denoising}

A graph structure needs to be defined before setting and applying graph based approaches for image denoising. In one example, a 4-connected simple graph structure is as shown in Fig.~\ref{fig:graph} (a). Each pixel is connected only to its four immediate neighbors in the pixel space. All pixels within the image or an image slice share a single graph and would be processed within one graph spectral domain. In Section \ref{ssec:pgs}, we propose an adapted version of the graph structure in Fig.~\ref{fig:graph} (b) to handle 3D image denoising with help from depth information.

After the connection structure in the graph is defined, a weight needs to be assigned for each graph edge. A well-known approach is to assign bilateral weights such as in \cite{gadde2013bilateral} and \cite{tian_icip2014}, where the weights $w_{ij}$ are defined by,

\begin{equation}
  w_{ij}=\exp(-\frac{\left\Vert p_{i}-p_{j}\right\Vert ^{2}}{2\sigma_{s}^{2}})\exp(-\frac{\left(x_{in}\left[i\right]-x_{in}\left[j\right]\right)^{2}}{2\sigma_{r}^{2}}).
\label{eq:weights_BF}
\end{equation}

The first exponential term is a spatial distance penalty ($p_{i}$ refers to the pixel's spatial location) and the second exponential term is an intensity distance penalty ($x_{in}$ refers to the intensity value from a guide image). We note that for the 4-connected simple graph structure as shown in Fig.~\ref{fig:graph} (a) the first exponential term is a constant. The guide image is the image itself in \cite{gadde2013bilateral} and a warped image in \cite{tian_icip2014}. Once the joint bilateral graph is constructed, the underlying filter as defined in (\ref{eq:image_filter_matrix}) is referenced as Joint Bilateral Filter (JBF) hereinafter. 

%
%

\section{Proposed Chebyshev and CG Graph Filtering}
\label{sec:ppsgf}

\subsection{Proposed graph structure}
\label{ssec:pgs}

We use a graph structure similar to the 4-connected graph shown in Fig.~\ref{fig:graph}(b). The edge weights of the links are defined as in \eqref{eq:weights_BF} with guided image being the warped image. A standard depth image based rendering (DIBR) process is employed to generate the guidance image as described below.

For each pixel location $i=[u,v]$ in the current view, a corresponding location $i'=[u',v']$ can be determined based on the camera parameters and the pin hole camera model. Hence, $x_{in}(i)$ in the warped image is obtained from $x'_{in}(i')$ in the other high quality view. Note that although $[u,v]$ is always at a pixel at integer location, $[u',v']$ may point to a subpixel location. As a result, subpixel interpolation in the high quality view needs be performed to maintain accuracy. In this paper, we use the 8- or 7- tap interpolation filter defined in H.265/HEVC video coding standard \cite{sullivanhevc}.

Occlusions may occur near foreground objects during the warping process, which are marked as hole areas. A typical hole filling algorithm would propagate the background pixels into the holes using inpainting algorithms. Since the filled holes are often unreliable, we do not use the filled pixels as guidance to construct the graph. Instead, we propose to simply avoid any links to or from a hole pixel. As shown in Fig.~\ref{fig:graph}~(b), as there is a hole pixel below pixel $p_i$, there are three links left for this pixel. More complex graph connections are subject to future research.

A conventional filter corresponding to the underlying filter of the above graph is herein referenced as JBF, since joint bilateral weights are used.

\subsection{Graph filtering via subspace projection}

The graph filtering problem can be viewed in the context of a general task of applying a function $h(\mathbf{L})$ of the graph Laplacian $\mathbf{L}$ to an input signal ${\mathbf{x}}_{in}$ thereafter for brevity denoted by $\mathbf{b}$, such that,

\begin{equation}\label{eq:generalFilter}
  \mathbf{x}^* = h(\mathbf{L}) \mathbf{b}.
\end{equation}
For image denoising, the goal is to suppress high frequency noise in which case the graph filter $h(\mathbf{L})$ is a low pass filter.

For general functions $h(\mathbf{L})$, e.g., such as the heat kernel $e^{-t\mathbf{L}}$, it becomes intractable to exactly compute the action of $h(\mathbf{L})$ on $\mathbf{b}$ via the eigendecomposition of $\mathbf{L}$ if the dimensionality of $\mathbf{b}$ is large. To overcome this difficulty, we propose instead to setup and evaluate the projection of $\mathbf{x}^*$ onto an appropriate low dimensional subspace $\mathcal{K}$. Denoting by $\mathbf{P}_{\mathcal{K}}$ the projector onto the subspace $\mathcal{K}$, the projection $\mathbf{x}^*_{\mathcal{K}}$ of the solution $\mathbf{x}^*$ of \eqref{eq:generalFilter} in the subspace $\mathcal{K}$ is given by,

\begin{equation}\label{eq:generalFilterProjected}
  \mathbf{x}^*_{\mathcal{K}} = \mathbf{P}_{\mathcal{K}}h(\mathbf{L})\mathbf{b}.
\end{equation}

If the subspace $\mathcal{K}$ is $\mathbf{L}$-invariant, i.e., $\mathbf{L}\mathcal{K}\subseteq\mathcal{K}$ then  
$  \mathbf{P}_{\mathcal{K}}h(\mathbf{L}) = h(\mathbf{P}_{\mathcal{K}}\mathbf{L}\mathbf{P}_{\mathcal{K}}),$
and thus $\mathbf{P}_{\mathcal{K}}h(\mathbf{L})\mathbf{b}$ can be efficiently computed for an arbitrary function $h(\cdot)$, e.g., utilizing a basis of the  subspace $\mathcal{K}$ made of eigenvectors of $\mathbf{L}.$


The optimal choice of the  $\mathbf{L}$-invariant subspace $\mathcal{K}$ is to apply a principal component analysis to the matrix $h(\mathbf{L})$ and select the subspace  $\mathcal{K}$ spanned by the principal eigenvectors, corresponding to the largest by absolute value eigenvalues of $h(\mathbf{L})$. This choice implies that  $\mathbf{P}_{\mathcal{K}}h(\mathbf{L}) \approx h(\mathbf{L})$ is the best small-rank approximation. Since the matrices $h(\mathbf{L})$ and $\mathbf{L}$ share the same eigenvectors, one can actually compute eigenvectors of $\mathbf{L}$ corresponding to the largest by absolute value eigenvalues of $h(\mathbf{L})$. If the shape of the function  $h(\cdot)$ is known in advance, one may be able to select the regions of interest in the spectrum of $\mathbf{L}$ for eigenvector computations. We note that this construction of the subspace  $\mathcal{K}$ does not explicitly depend on ${\mathbf{x}}_{in}=\mathbf{b}$. It may offer a dramatic computational cost improvement, compared to the full eigenvalue decomposition of $h(\mathbf{L})$ or $\mathbf{L}$, if an efficient iterative method is used to compute the basis of the subspace  $\mathcal{K}$. 

In this paper, we are interested specifically in low-pass filters, so we concentrate on using the Krylov subspace projection technique \cite{GraphSP_HeatKernel:2008,Krylov:1997}, i.e.,\ we approximate the solution $\mathbf{x}^*$ by a vector in the order-$(k+1)$ Krylov subspace,
\begin{equation}\label{eq:Krylov}
  \mathcal{K} = \textrm{span}\{\mathbf{b}, \mathbf{Lb}, \dots, \mathbf{L}^{k}\mathbf{b}\}.
\end{equation}

The Krylov subspace is known to approximate well eigenvectors corresponding to extreme eigenvalues, thus it should be an appropriate choice for high- and low-pass filters. Although  Krylov subspace \eqref{eq:Krylov}  is not normally exactly $\mathbf{L}$-invariant, we can still attempt using  $h(\mathbf{P}_{\mathcal{K}}\mathbf{L}\mathbf{P}_{\mathcal{K}})$
as the only practically available, within the constraints of polynomial filtering, replacement for  $\mathbf{P}_{\mathcal{K}}h(\mathbf{L}).$
 Moreover, Krylov subspace \eqref{eq:Krylov} is built starting with the initial image ${\mathbf{x}}_{in}=\mathbf{b}$, therefore it only takes into account the  $\mathbf{L}$-spectral modes actually present in the image. 

An implementation of a filter using the projection on the degree $k$ Krylov subspace can be especially simple if the filter function $h(\cdot)$ itself is a polynomial of degree $k$ or smaller,  $h(\lambda) = p_{k}(\lambda)$, since 
$\mathbf{x}^*_{\mathcal{K}} = \mathbf{P}_{\mathcal{K}} p_{k}(\mathbf{L})\mathbf{b}
= p_{k}(\mathbf{L})\mathbf{b},$
where the equality holds since the subspace $\mathcal{K}$ contains all possible vectors $p_{k}(\mathbf{L})\mathbf{b}$ for any polynomial $p_{k}(\cdot)$ of degree $k$ or smaller, i.e. $p_{k}(\mathbf{L})\mathbf{b} \in \mathcal{K}$.
Computational gain is achieved when the matrix $\mathbf{L}$ is sparse and the degree $k$ is small. 

\subsection{Chebyshev polynomial graph spectral denoising}

In the graph-based image denoising problem, we wish to apply a low pass filter to the graph spectrum of the noisy image in order to remove high frequency noise. Restricting ourselves to polynomial filtering in this work, e.g., due to computational considerations, we can setup a problem of designing optimal polynomial low pass filters. Our first optimal polynomial low pass filter is based on Chebyshev polynomials. 

Specifically, we propose to use a degree $k$ Chebyshev polynomial $h_{\text{k-CHEB}}$ defined over the interval $[0, 2]$ with a stop band extending from $l \in (0,2)$ to 2. Since we define the graph spectrum in terms of the eigenspace of the symmetric normalized Laplacian $\mathbf{L}$, all the eigenvalues of $\mathbf{L}$ lie in the interval $[0, 2]$. 
The construction of a Chebyshev polynomial is easily obtained by computing the roots of the degree $k$ Chebyshev polynomial 
$\hat{r}(i) = \cos\left(
{\pi(2i-1)}/{2k}\right) \ \text{for} \ i = 1\dots k$. 
over the interval $[-1, 1]$, then shifting the roots to the interval $[l, 2]$ via linear transformation to obtain the roots $r_i$ of the polynomial $h_{\text{k-CHEB}}$, and scaling the polynomial using $r_0$ such that $h_{\text{k-CHEB}}(0) = 1$. 
This results in formula,
\[
h_{\text{k-CHEB}}(\lambda) = p_{k}(\lambda) = \prod_{i=1}^{k}\left(1-\frac{\lambda}{r_i}\right) = r_0 \prod_{i=1}^{k}\left(r_i - \lambda\right);
\] and we can compute $\mathbf{x}^*_{\mathcal{K}}$ by evaluating $\mathbf{x}^{i}=r_i\mathbf{x}^{i-1} - \mathbf{L}\mathbf{x}^{i-1}$ iteratively for $i=1,\ldots,k,$ where $\mathbf{x}^{0}=r_0 \mathbf{b}.$

Chebyshev polynomials are minimax optimal, uniformly suppressing all spectral components on the  interval $[l, 2]$ and growing faster than any other polynomial outside of $[l, 2]$. The stop band frequency $l$ remains a design parameter that needs to be set prior to filtering. In the next subsection, we propose a variational parameter-free method.

\subsection{Conjugate Gradient method for Krylov denoising}
\label{ssec:cggf}

Since $\mathbf{L}$ is a high-pass filter, its Moore-Penrose pseudoinverse $\mathbf{L}^\dagger$ is a low-pass filter, so we propose designing a low-pass graph spectral filter by choosing $h(\mathbf{L}) = \mathbf{L}^\dagger$, i.e.,\ setting $\mathbf{x}^*= \mathbf{L}^\dagger \mathbf{b}$. However, in contrast to the  approach of \cite{gadde2013bilateral}, where $h_{\text{GBJBF}}(\cdot)$ is approximated by a truncated Chebyshev polynomial expansion, we do not attempt to approximate the inverse function by a polynomial. Instead, we use the subspace projection technique, formulating the graph filtering problem as a constrained quadratic program,
\begin{equation}\label{eq:graphFilter_Optimization}
\begin{array}{l}
	\mathbf{x}^*_{\mathcal{\bar{K}}} = \arg\min\limits_{\mathbf{x} \in \mathcal{\bar{K}}} \quad \mathbf{x}^T\mathbf{L x} - 2\mathbf{x}^T\mathbf{f},\\
	\mathcal{\bar{K}} = \mathbf{x}^0 + \textrm{span} \left\{ \mathbf{f-Lx^0}, \ldots, \mathbf{L}^k\mathbf{(f-Lx^0)} \right\},
\end{array}
\end{equation}
where the  initial approximation $\mathbf{x}^0$ and the vector  $ \mathbf{f}$ remain to be chosen. 
The solution to problem \eqref{eq:graphFilter_Optimization} can be computed by running $k$ iterations of the CG method \cite{Hestenes&Stiefel:1952}. 
We denote by $h_{\text{k-CG}}$ the $k$-step CG filter with $\mathbf{x}^0=\mathbf{f}=\mathbf{b}$, in which case  $\mathcal{\bar{K}} \subseteq \mathcal{K}$---the order-$(k+1)$ Krylov subspace defined in \eqref{eq:Krylov}. We denote by $h_{\text{k-CG0}}$ the $k$-step CG filter with $\mathbf{x}^0=\mathbf{b}$ and $\mathbf{f}=0$. Yet another choice $\mathbf{x}^0=0$ and $\mathbf{f}=\mathbf{b}$ results in  $\mathcal{\bar{K}}$ equal to the order-$k$ Krylov subspace $\textrm{span}\{\mathbf{b}, \mathbf{Lb}, \dots, \mathbf{L}^{k-1}\mathbf{b}\}$.

\begin{figure}
\begin{minipage}[b]{.48\linewidth}
  \centering
  \centerline{\includegraphics[width=4.0cm]{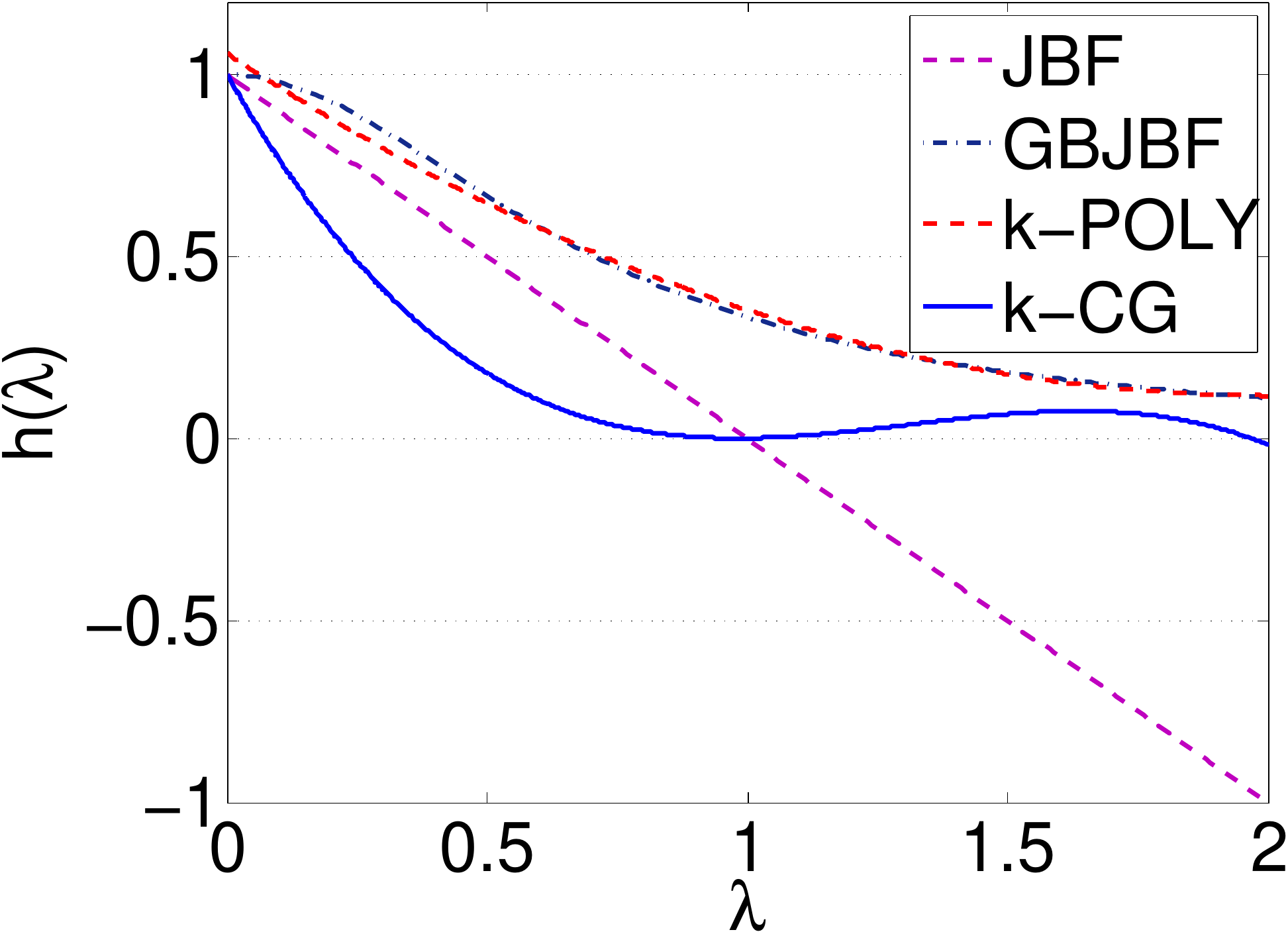}} 
  \centerline{(a)}
\end{minipage}
\hfill
\begin{minipage}[b]{.48\linewidth}
  \centering
  \centerline{\includegraphics[width=4.0cm]{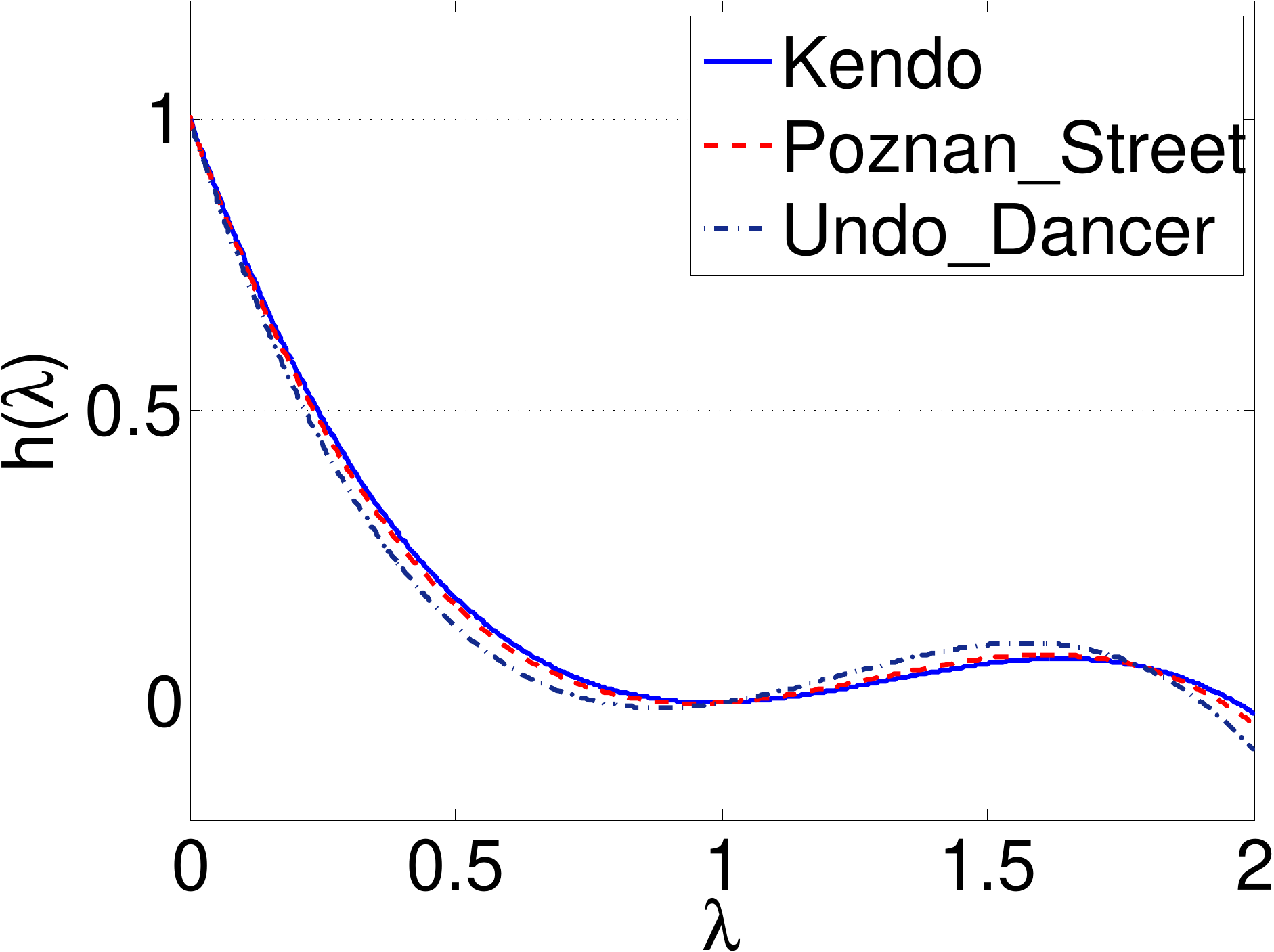}} 
  \centerline{(b)}
\end{minipage}
\caption{Spectral responses of the graph filters. (a) Different graph filters: JBF, GBJBF, k-POLY \cite{gadde2013bilateral}, and proposed k-CG; (b) Proposed CG filter with different sequences: Kendo, Pozan\_Street, and Undo\_Dancer.}
\label{fig:specall}
\end{figure}
\begin{figure}
\begin{minipage}[b]{.48\linewidth}
  \centering
  \centerline{\includegraphics[width=4.0cm]{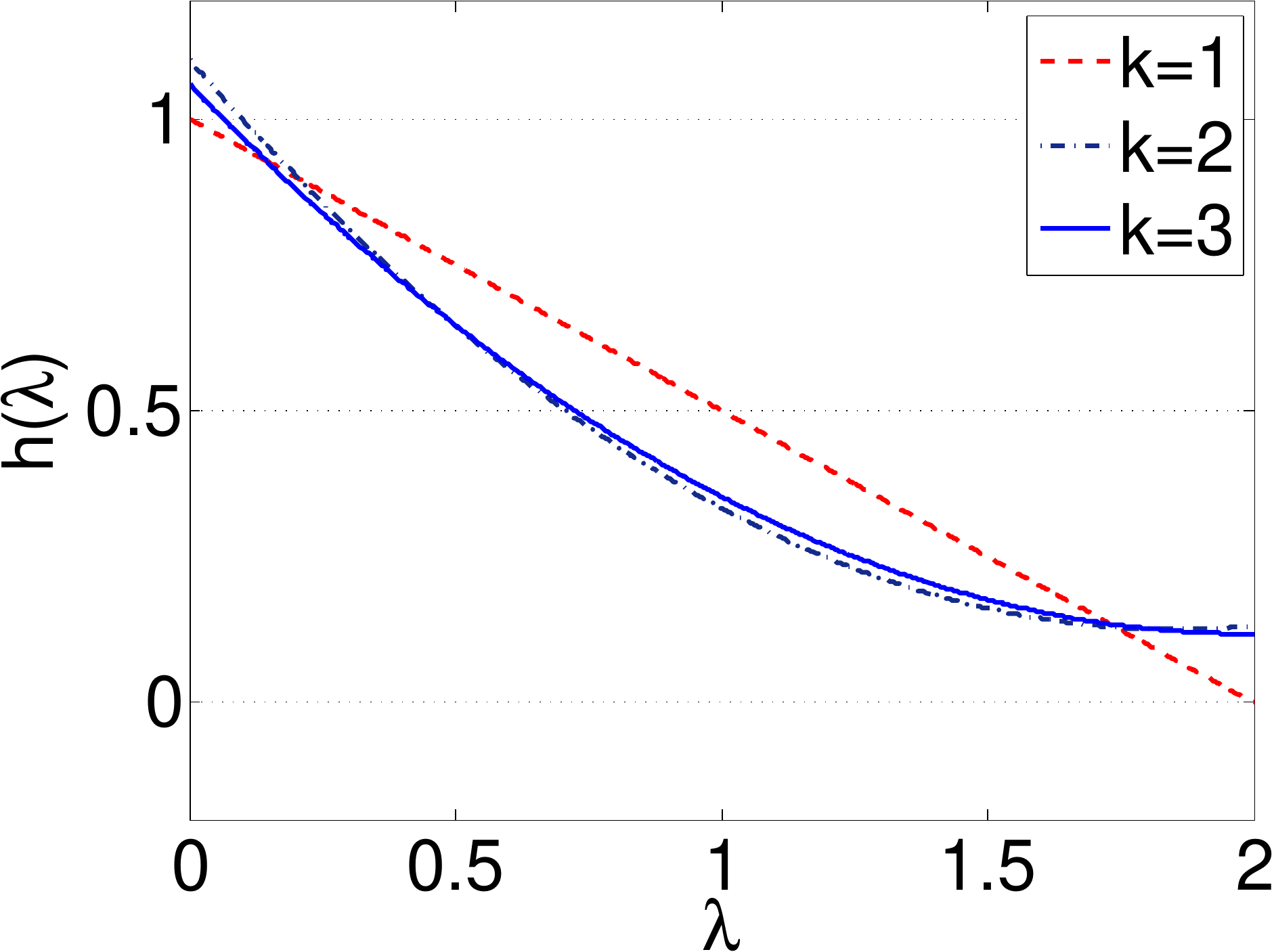}} 
  \centerline{(a)}
\end{minipage}
\hfill
\begin{minipage}[b]{0.48\linewidth}
  \centering
  \centerline{\includegraphics[width=4.0cm]{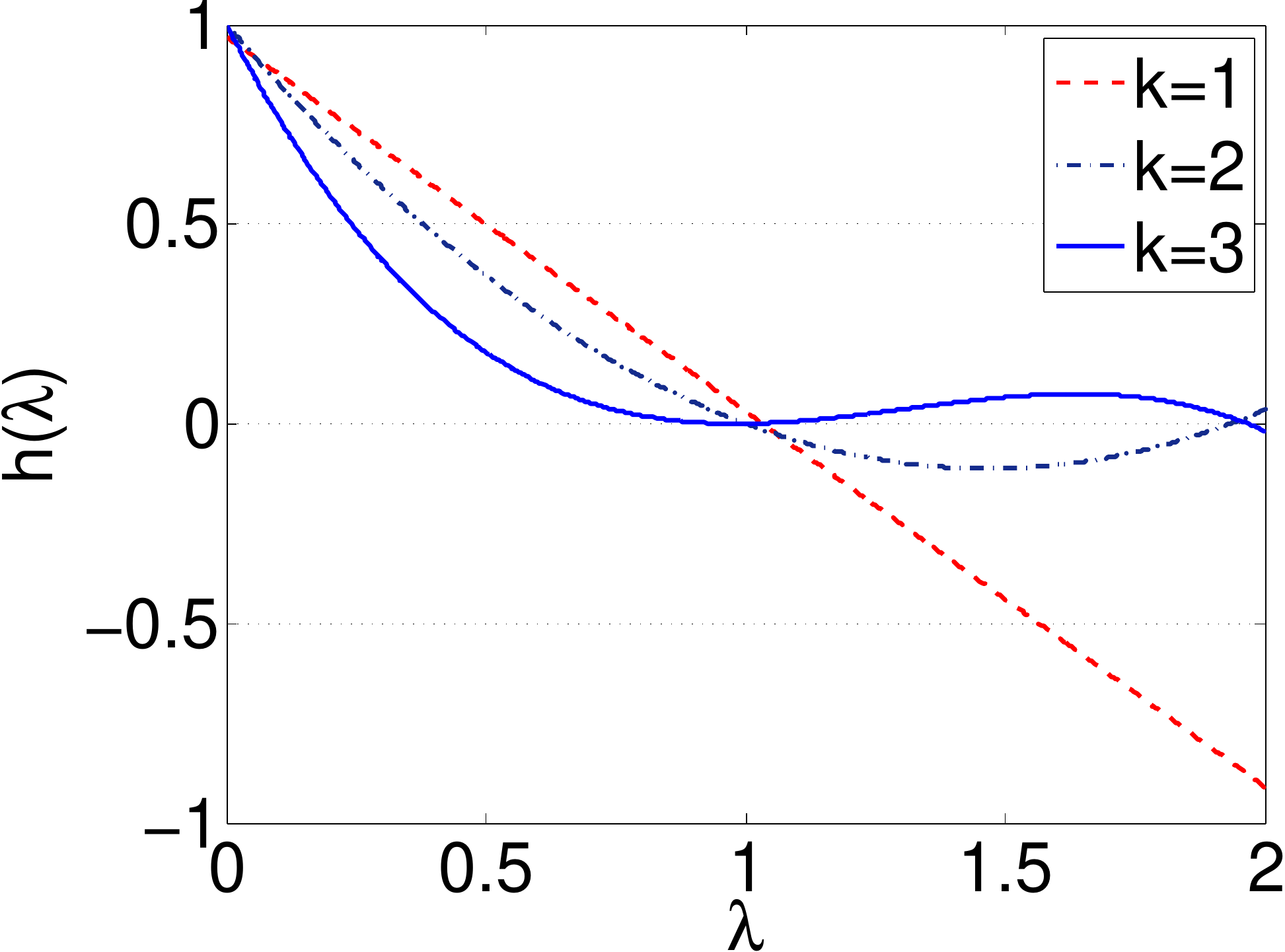}}
  \centerline{(b)}
\end{minipage}
\caption{Spectral response with at different iterations, (a) k-POLY \cite{gadde2013bilateral}; (b) k-CG in this paper.}
\label{fig:speciter}
\end{figure}

Fig.~\ref{fig:specall} (a) compares the spectral response $h(\lambda)$ of three iterations of the proposed 3-CG approach with the benchmark methods: JBF, GBJBF, and 3-POLY---a degree-3 polynomial $h_{\text{3-POLY}}$ approximation of GBJBF as in \cite{gadde2013bilateral}. The CG algorithm adaptively adjusts the spectral response based on the input image as shown in Fig.~\ref{fig:specall}~(b). 
Fig.~\ref{fig:speciter}~(a) and (b) plot $h_{\text{k-POLY}}$ and $h_{\text{k-CG}}$ and illustrate the variation in spectral response over different iterations for k-POLY and k-CG with $k=1,\,2,\,3.$

\section{Experiments and Discussions}
\label{sec:ead}

We evaluate the performance of the proposed approaches for graph filtering on stereo plus depth 3D video denoising. In our experimental setup, one view is captured at high quality while the other view is degraded with Gaussian noise. The depth map associated with the stereo video is available. Three videos with varying characteristics are chosen: Kendo @ $1024\times768$ and Poznan\_Street are captured from natural scenes with estimated depth maps; Undo\_Dancer @ $1920\times1080$ is a synthetic sequence generated from 3D models with ground truth depth maps \cite{jct3vctc}. To filter pixels in the hole area, we simply apply a $3\times 3$ median filter.

\begin{table}
  \centering\caption{Denoised image PSNR results (dB)}
  \begin{tabular} {|l||c|c|c|}
    \hline
    Seqs & Kendo & Poznan\_Street & Undo\_Dancer \\
    \hline \hline
    JBF  & 32.53 & 32.78 & 31.90 \\
    \hline
    3-POLY & 32.28 & 33.05 & 31.97 \\
    \hline
    3-CG & 35.76 & 34.49 & 31.91 \\
    \hline
    3-CHEB & 35.67 & 34.35 & 31.92  \\
    \hline
  \end{tabular}

\label{fig:psnr}
\end{table}

\begin{figure}[t]

\begin{minipage}[b]{.48\linewidth}
  \centering
  \centerline{\includegraphics[width=4.0cm]{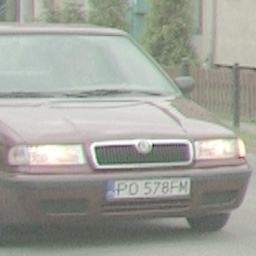}}
  \centerline{(a) Original Image.}\medskip
\end{minipage}
\hfill
\begin{minipage}[b]{0.48\linewidth}
  \centering
  \centerline{\includegraphics[width=4.0cm]{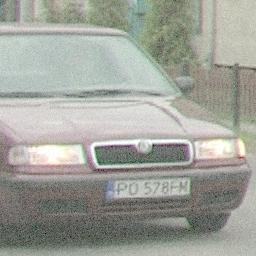}}
  \centerline{(b) Noisy Image.}\medskip
\end{minipage}

\begin{minipage}[b]{.48\linewidth}
  \centering
  \centerline{\includegraphics[width=4.0cm]{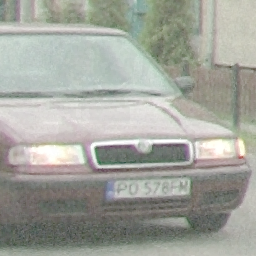}}
  \centerline{(c) Filtered Image, 3-POLY.}\medskip
\end{minipage}
\hfill
\begin{minipage}[b]{0.48\linewidth}
  \centering
  \centerline{\includegraphics[width=4.0cm]{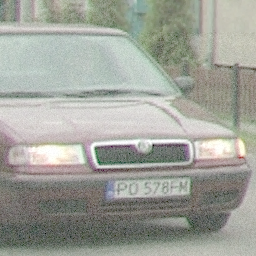}}
  \centerline{(d) Filtered Image, 3-CG.}\medskip
\end{minipage}

\caption{Comparison of filtered images.}
\label{fig:street}
\end{figure}

\subsection{Performance evaluations}

We compare the denoising performance of four approaches using the same graph structure as proposed in Section \ref{ssec:pgs}. The first benchmark method is JBF, which shares the same weights as the underlying graph. The second benchmark method is 3-POLY, which is a degree-3 polynomial approximation to the regularization solution \eqref{eq:gbjbf_vertex} of GBJBF with parameter $\rho = 2$. We then run our degree-3 Chebyshev filter (3-CHEB) and 3 iterations of our CG approach (3-CG) and compare the denoised image to the benchmark techniques. 

The denoised image PSNR results are summarized in Table~\ref{fig:psnr}. On the one hand, all methods demonstrate similar denoising performance for the Undo\_Dancer sequence, where the PSNR differences vary between 0.01 - 0.07 dB. This result might be explained by the lack of complex texture in this synthetic sequence. 

On the other hand, we observe that our CG approach results in a significant improvement of 3.23 dB and 1.44 dB in PSNR over the best benchmark techniques for sequences Kendo and Poznan\_Street, respectively. The quality improvement is also verified in subjective evaluations. Fig.~\ref{fig:street} (c) and (d) are the filtered images from 3-POLY and 3-CG. Notice that the CG approach preserves texture details better than 3-POLY, e.g.,\ plate numbers resulted from the CG approach are easier to recognize. The content adaptation feature of CG is illustrated by various spectral response curves for different sequences in Fig.~\ref{fig:specall} (b).

\begin{figure}[t]
\centering
\begin{minipage}[b]{.55\linewidth}
  \centering
  \centerline{\includegraphics[width=4.0cm]{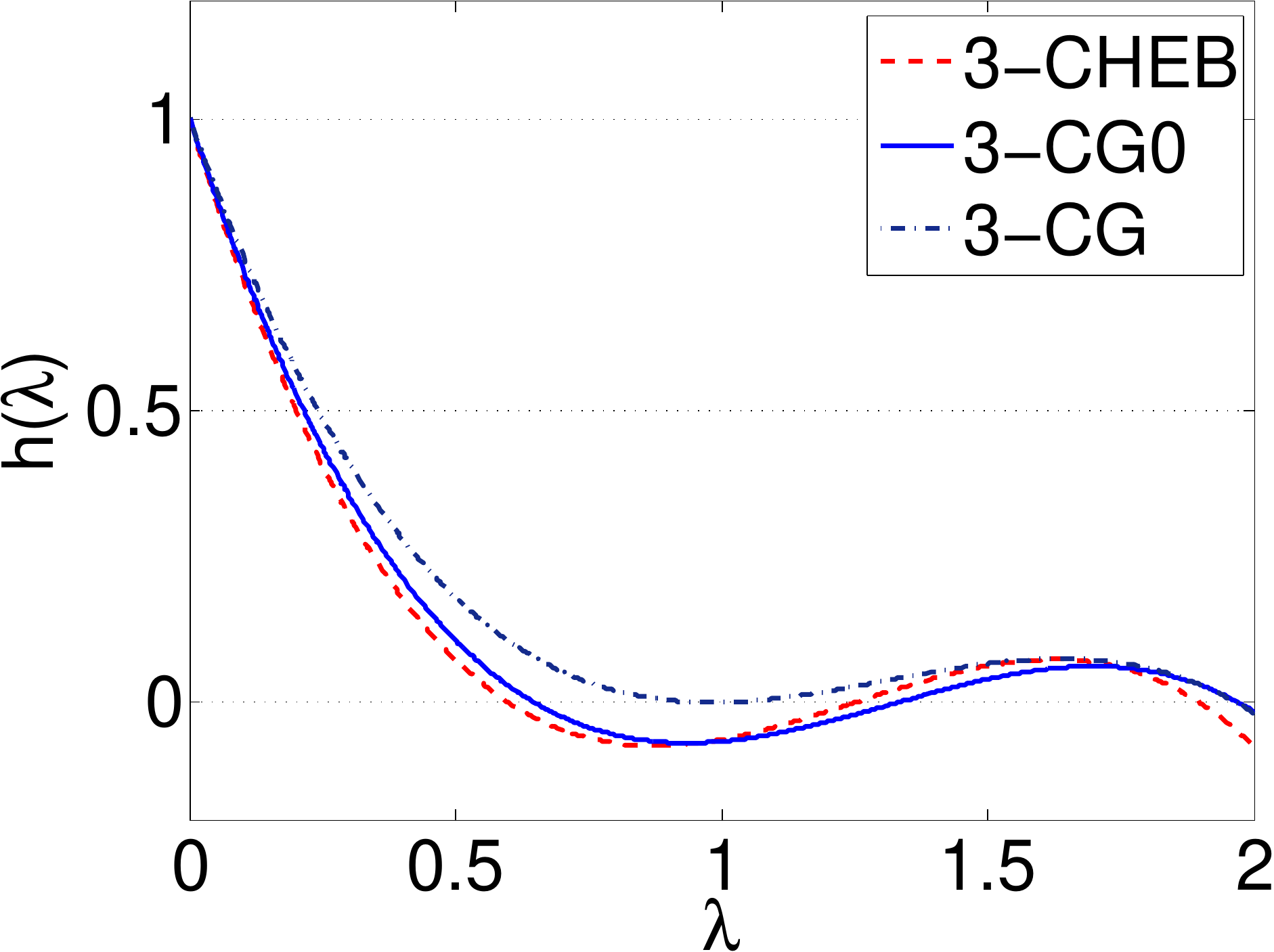}}
\end{minipage}
\caption{Comparison of spectral response of the proposed 3-CHEB and 3-CG filters.}
\label{fig:cheb_cg}
\end{figure}

The Chebyshev filter that we design has a stop band at $l = 0.5$. Notice that the denoised image PSNR is almost identical to the CG PSNR. That can be explained by our choice of $l$, which is close to the effective stop band of our CG filter k-CG0. We plot examples of the spectral response of our k-CG,  k-CG0, and k-CHEB filters in Fig.~\ref{fig:cheb_cg}.

\subsection{Patch-based processing}

In order to reduce memory requirements in implementing the above filtering schemes, we use $64\times64$ disjoint patches of the image that are filtered independently. Moreover, the patch based approach is favorable for parallel implementation and could be synchronized with the coding unit structures used in a typical video coding framework such as H.265/HEVC. 

\begin{figure}[t]
\begin{minipage}[b]{.48\linewidth}
  \centering
  \centerline{\includegraphics[width=4.0cm]{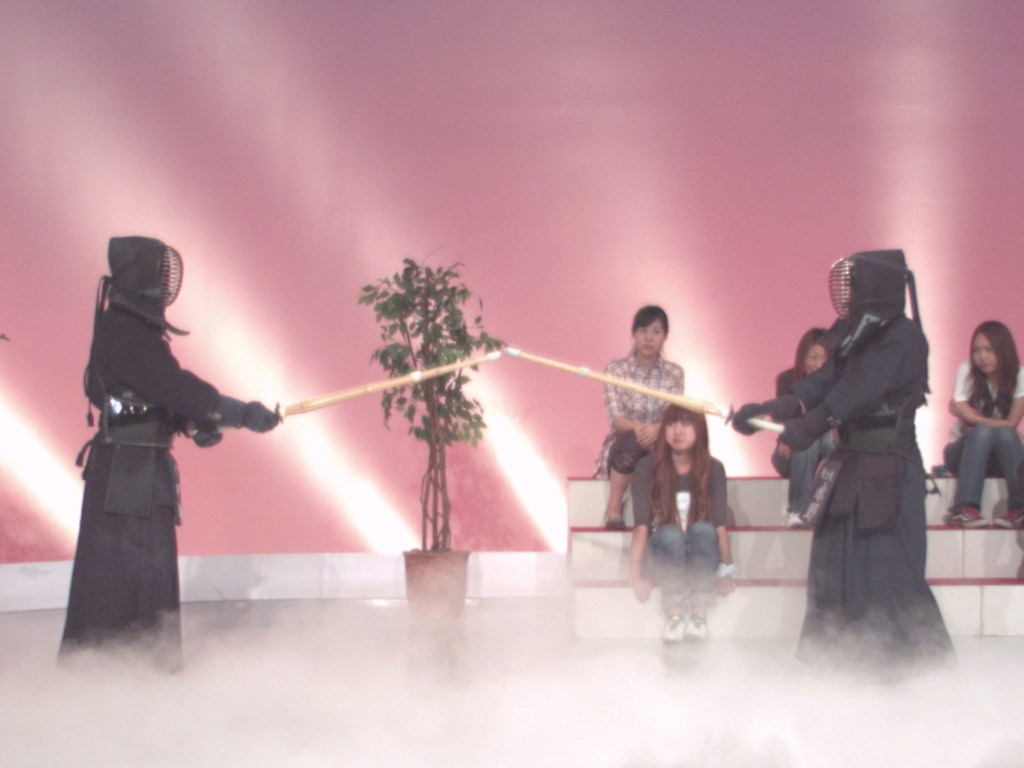}}
  \centerline{(a) Original Image.}
\end{minipage}
\hfill
\begin{minipage}[b]{0.48\linewidth}
  \centering
  \centerline{\includegraphics[width=4.0cm]{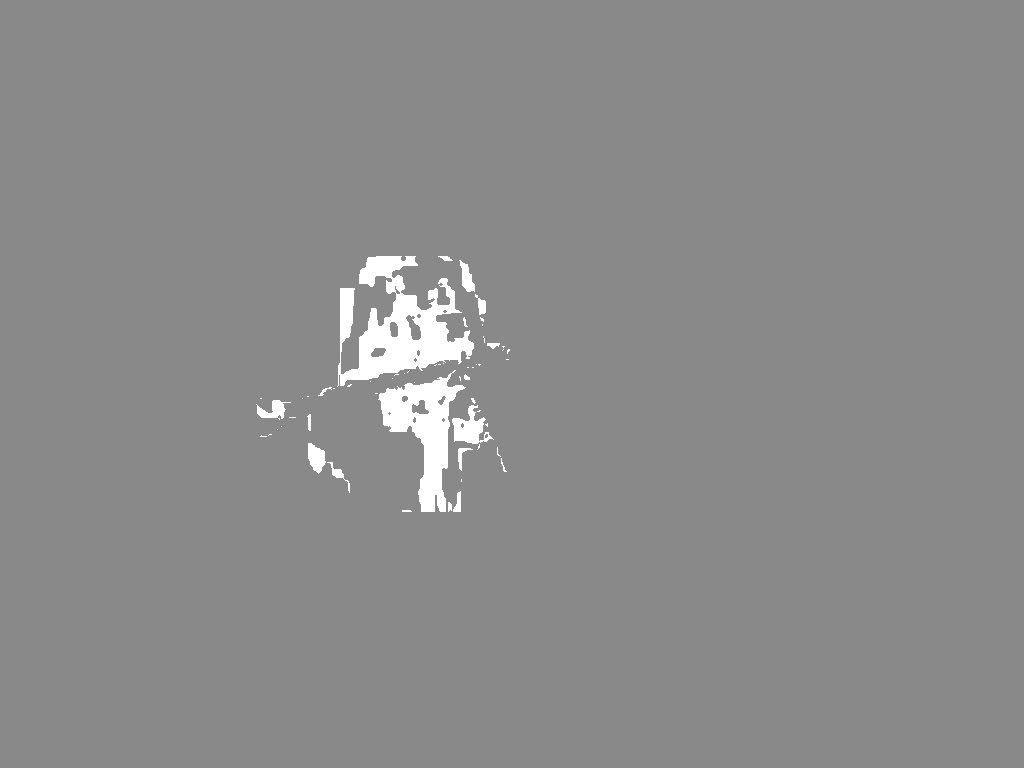}}
  \centerline{(b) Mask Map.}
\end{minipage}

\begin{minipage}[b]{.48\linewidth}
  \centering
  \centerline{\includegraphics[width=4.0cm]{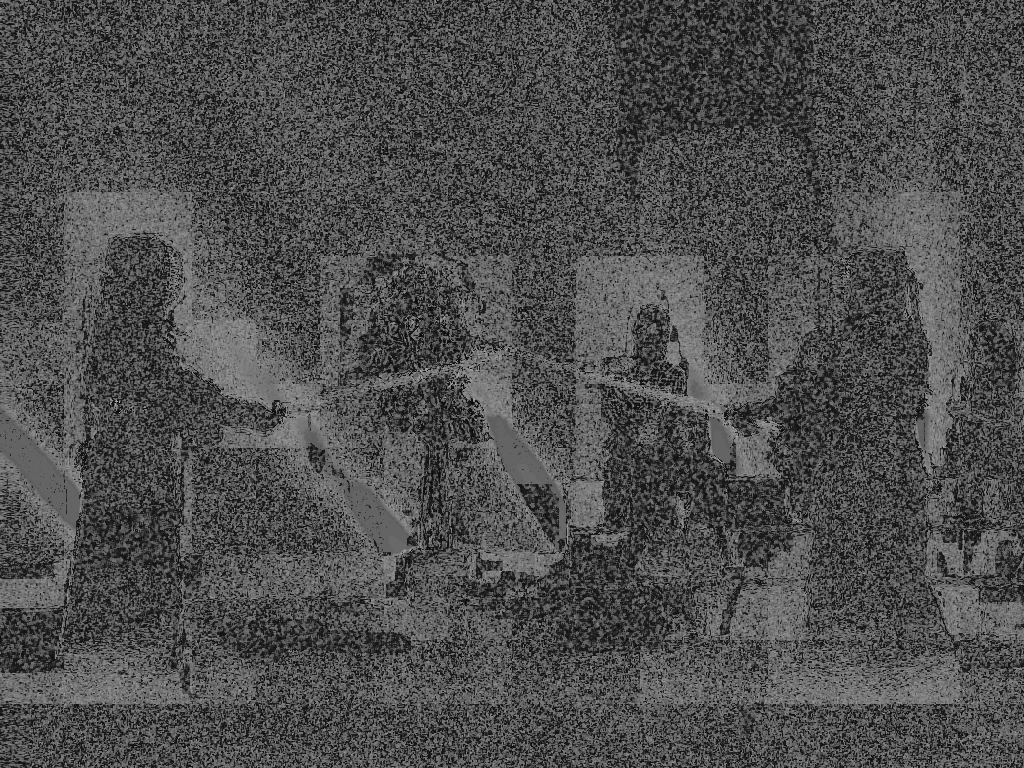}}
  \centerline{(c) Diff. Image, 3-POLY.}
\end{minipage}
\hfill
\begin{minipage}[b]{0.48\linewidth}
  \centering
  \centerline{\includegraphics[width=4.0cm]{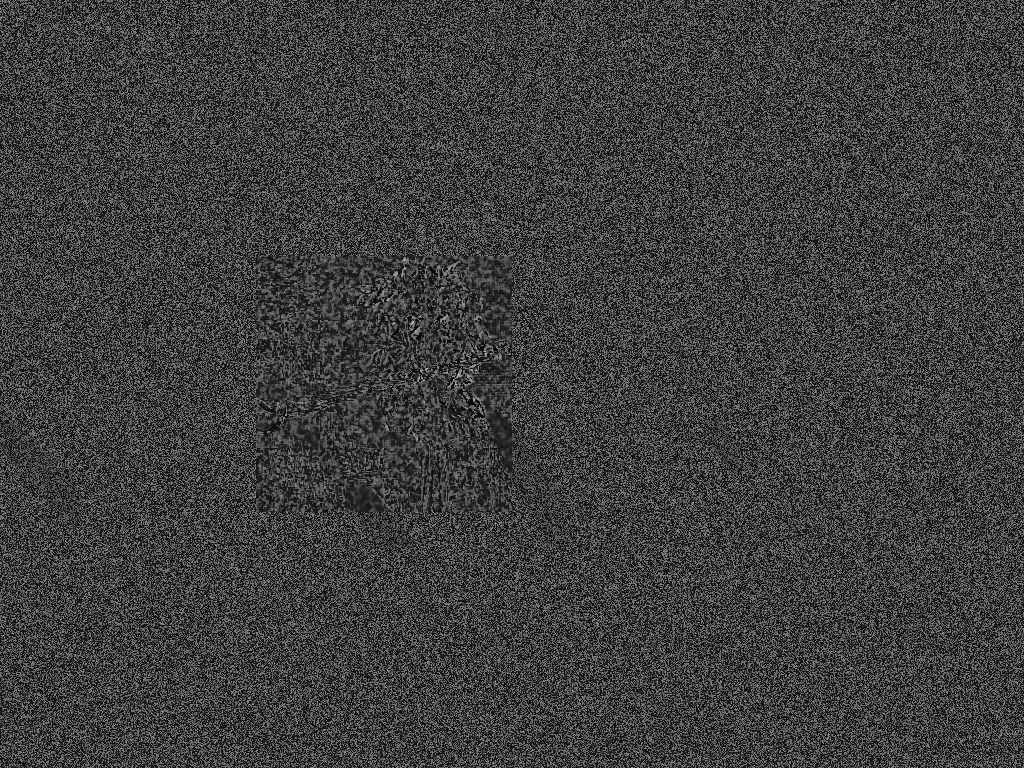}}
  \centerline{(d) Diff. Image, 3-CG.}
\end{minipage}

\caption{Block artifacts along patch boundaries.}
\label{fig:kendo}
\end{figure}

A subjective evaluation of the patch based processing approach shows that GBJBF sometimes exhibits block artifacts across patch boundaries. Fig.~\ref{fig:kendo} (c) illustrates these block artifacts as witnessed by the difference between the noisy and denoised images. On the other hand, our CG approach does not suffer from blocky denoising as shown in Fig.~\ref{fig:kendo} (d). The only block viewed in the difference image of the CG approach is due to the switch between the median filter applied to the hole pixels, shown in Fig.~\ref{fig:kendo} (b), and the CG technique.

Finally, the polynomial degree $k$, which we set equal to $3$ for $64\times64$ patches, tends to increase with larger patch sizes. We note that for k-POLY, the degree needs to be determined beforehand so as to determine the polynomial coefficients and cannot be incrementally increased without recalculating the coefficients and restarting the filtering process. 

In contrast, the k-CHEB approach allows incrementally increasing the degree $k,$ although only in such a way that the higher degree polynomials share the same roots as the previously computed lower degree polynomials, which is possible for some degrees which are powers of $2$ and $3$, or their products. Moreover, the k-CG technique allows for a seamless incremental increase in the degree of the polynomial by simply increasing the number of iterations.

\section{Conclusions and Future Work}

We propose Chebyshev and conjugate gradient filters for graph based image denoising. Our approach performs graph-based filtering on the noisy image by directly computing the projection of the desired filtered image onto a lower dimensional Krylov subspace for the graph Laplacian using novel Chebyshev or conjugate gradient polynomial graph filters. We demonstrate through numerical simulations that our proposed technique produces subjectively cleaner images with about 1-3 dB improvement in PSNR over existing polynomial graph filters. The proposed approaches are not limited to 3D image denoising and could be used for improving other types of images, provided that guidance images are constructed under proper contexts, which is subject to future work.

\bibliographystyle{IEEEbib}
\bibliography{refs}  

\begin{thebibliography}{10}

\bibitem{shuman_signal_2013}
D.I. Shuman, S.K. Narang, P.~Frossard, A.~Ortega, and P.~Vandergheynst,
\newblock ``The emerging field of signal processing on graphs: Extending
  high-dimensional data analysis to networks and other irregular domains,''
\newblock {\em Signal Processing Magazine, IEEE}, vol. 30, no. 3, pp. 83--98,
  2013.

\bibitem{Sunil_GlobalSip13}
S.~K. Narang, A.~Gadde, E.~Sanou, and A.~Ortega,
\newblock ``Localized iterative methods for interpolation in graph structured
  data,''
\newblock in {\em Signal and Information Processing (GlobalSip), 1st IEEE
  Global Conf.}, Dec. 2013.

\bibitem{gadde2013bilateral}
A.~Gadde, S.~K. Narang, and A.~Ortega,
\newblock ``Bilateral filter: Graph spectral interpretation and extensions,''
\newblock {\em ICIP 2013}, October 2013.

\bibitem{wang_icassp14}
Y.~Wang, A.~Ortega, D.~Tian, and A.~Vetro,
\newblock ``A graph-based joint bilateral approach for depth enhancements,''
\newblock in {\em to appear in ICASSP 2014}, 2014.

\bibitem{tian_icip2014}
D.~Tian, H.~Mansour, A.~Vetro, Y.~Wang, and A.~Ortega,
\newblock ``Depth-assisted stereo video enhancement using graph-based
  approaches,''
\newblock in {\em submitted to ICIP 2014}, 2014.

\bibitem{GraphSP_HeatKernel:2008}
F.~Zhang and E.~R. Hancock,
\newblock ``Graph spectral image smoothing using the heat kernel,''
\newblock {\em Pattern Recogn.}, vol. 41, no. 11, pp. 3328--3342, Nov. 2008.

\bibitem{WaveletsGraphs:2011}
D.~K. Hammond, P.~Vandergheynst, and R.~Gribonval,
\newblock ``Wavelets on graphs via spectral graph theory,''
\newblock {\em Applied and {C}omputational {H}armonic {A}nalysis}, vol. 30, no.
  2, pp. 129--150, 2011.

\bibitem{leo_book}
D.~K. Hammond, L.~Jacques, and P.~Vandergheynst,
\newblock ``Image denoising with nonlocal spectral graph wavelets,''
\newblock in {\em Image Processing and Analysis with Graphs}, O.~Lezoray and
  L.~Grady, Eds., pp. 207--236. CRC Press, 2012.

\bibitem{sullivanhevc}
G.J. Sullivan, J.~Ohm, Woo-Jin Han, and T.~Wiegand,
\newblock ``Overview of the high efficiency video coding {(HEVC)} standard,''
\newblock {\em Circuits and Systems for Video Technology, IEEE Transactions
  on}, vol. 22, no. 12, Dec 2012.

\bibitem{Krylov:1997}
M.~Hochbruck and C.~C.~Lubich,
\newblock ``{On Krylov subspace approximations to the matrix exponential
  operator},''
\newblock {\em SIAM Journal on Numerical Analysis}, vol. 34, no. 5, pp.
  1911--1925, 1997.

\bibitem{Hestenes&Stiefel:1952}
M.~R. Hestenes and E.~Stiefel,
\newblock ``Methods of conjugate gradients for solving linear systems,''
\newblock {\em Journal of research of the National Bureau of Standards}, vol.
  49, pp. 409--436, 1952.

\bibitem{jct3vctc}
K.~Muller and A.~Vetro,
\newblock ``Common test conditions of {3DV} core experiments,''
\newblock in {\em JCT3V meeting, JCT3V-G1100}, January 2014.

\end{thebibliography}

\end{document}